\documentclass[preprint,12pt,authoryear]{elsarticle}
\usepackage{graphicx}
\usepackage{multirow}
\usepackage{amsmath,amssymb,amsfonts}
\usepackage{booktabs}
\usepackage{float}
\usepackage{xcolor}
\usepackage[hidelinks]{hyperref}
\usepackage{xurl}
\usepackage[caption=false]{subfig}
\usepackage{url}
\graphicspath{{Figures/}}
\journal{International Journal of Information Management Data Insights}

\begin{document}
\sloppy

%%%%%%%%%%%%%%%%%%%%%%%%
\begin{frontmatter}

\title{BanClickThumb: A Multimodal Dataset and Transformer Fusion Benchmarks for Clickbait Detection in Bengali YouTube Videos}

\author[1]{Md. Ariful Islam\corref{cor1}}
\author[1]{Md. Tanvirul Islam}
\author[1]{Md. Maruf Hossain Miru}
\author[1]{Md Khalid Syfullah}
\cortext[cor1]{Corresponding author.}
\affiliation[1]{organization={Department of Computer Science and Engineering, Bangladesh Army University of Science and Technology}, addressline={Saidpur Cantonment}, city={Saidpur}, country={Bangladesh}}

\begin{abstract}
Clickbait, where video titles and thumbnails exaggerate or misrepresent content, reduces user trust, wastes attention, and promotes misinformation on video-sharing platforms. Detecting Bengali clickbait remains challenging because publicly available multimodal datasets are limited. To address this gap, we introduce \textbf{BanClickThumb}, a curated dataset of 7,147 Bengali YouTube thumbnail-title pairs from five content domains, annotated by ten annotators with high agreement (Cohen's Kappa: 0.83--0.93). Using this dataset, we benchmark text-only, image-only, and multimodal approaches. Among unimodal models, \textbf{BanClickTextFormer} (XLM-RoBERTa) achieves 0.82 accuracy, while \textbf{BanClickImageFormer} (SwiftFormer) reaches 0.68. Our proposed multimodal model, \textbf{BanClickFusionFormer}, combines ViT and XLM-RoBERTa through intermediate fusion and achieves the best accuracy of 0.84. Error analysis shows that dense thumbnail text, figurative language, and culturally specific slang remain challenging. Our findings demonstrate the effectiveness of multimodal fusion for Bengali clickbait detection and provide a publicly available benchmark to support future research on low-resource multimodal content analysis.
\end{abstract}

\begin{keyword}
Clickbait detection \sep Bengali \sep Multimodal learning \sep YouTube \sep Vision Transformer \sep XLM-RoBERTa \sep Under-resourced languages \sep Benchmark dataset
\end{keyword}

\end{frontmatter}

%%%%%%%%%%%%%%%%%%%%%%%%
%%%%%%%%%%%%%%%%%%%%%%%%
\section{Introduction}
\label{sec:intro}

The economics of digital media are, at their core, an economics of attention. Video-sharing platforms have become the dominant channel through which billions of people consume news, education, and entertainment, with YouTube alone serving more than 2.5 billion monthly active users \citep{youtube_stats}. On such platforms, the thumbnail and title are the primary, and often the only, signals a viewer sees before deciding whether to click, making them the principal battleground for audience attention \citep{organic_reach_youtube}. Because creator revenue is directly coupled to views, a strong incentive exists to engineer these signals for maximum click-through rather than for faithful representation of content.

Clickbait exploits precisely this incentive. Defined as content ``designed to make readers want to click on a hyperlink, especially when the link leads to content of dubious value'' \citep{merriam_clickbait}, clickbait operates by opening a \emph{curiosity gap}: the title or thumbnail withholds or exaggerates information to provoke an urge that can only be resolved by clicking \citep{loewenstein1994psychology, blom2015click}. Beyond wasting user time, sustained exposure to deceptive framing degrades trust in online media and provides fertile ground for misinformation, whose psychological pull relies on many of the same attention-hijacking mechanisms \citep{Chen2015, pennycook2021psychology}. Automatic clickbait detection is therefore not merely a classification exercise; it is an information-quality problem with direct consequences for platform governance, advertising integrity, and public discourse.

A decade of research has produced increasingly capable detectors, from linguistically motivated classifiers over news headlines \citep{chakraborty2016stop, Potthast2016Corpus} to deep multimodal systems that jointly reason over titles, thumbnails, and audio \citep{baitradar2021, varshney2021, yu2024multimodal}. This progress is heavily skewed toward English. Bengali (Bangla), the seventh most spoken language in the world, remains severely under-resourced despite Bangladesh hosting one of the largest and fastest-growing YouTube audiences globally \citep{youtube_bd_users}. Existing Bengali resources are either text-only \citep{baitbuster2024, arfat2024bangla} or target adjacent tasks such as fake news detection \citep{faria2025multibanfakedetect}, and none provides annotated thumbnail-title pairs. This is a consequential gap: clickbait on YouTube is an inherently \emph{visual-linguistic} phenomenon, in which an exaggerated thumbnail and a sensational title jointly construct the deception. To the best of our knowledge, no publicly available, manually annotated multimodal dataset exists for Bengali YouTube clickbait detection.

To address this, we introduce \textbf{BanClickThumb}, a balanced multimodal dataset of 7,147 Bengali YouTube thumbnail-title pairs, and we establish transformer-based benchmarks across unimodal and multimodal settings. We develop three frameworks: \textbf{BanClickTextFormer}, a text-only detector over video titles; \textbf{BanClickImageFormer}, a vision-only detector over thumbnails; and \textbf{BanClickFusionFormer}, a multimodal architecture that systematically compares early, late, and intermediate fusion of the two modalities.

We investigate the following Research Questions (RQs):
\begin{itemize}
 \item \textbf{RQ1:} Can a large-scale, balanced, and reliably annotated multimodal dataset for Bengali YouTube clickbait be constructed, and how does it compare with existing clickbait resources?
 \item \textbf{RQ2:} How accurately do transformer-based text encoders and vision transformers detect Bengali clickbait when each modality is used in isolation, and which modality carries the stronger signal?
 \item \textbf{RQ3:} Does fusing titles and thumbnails improve detection over unimodal baselines, and at which stage of the network (early, late, or intermediate) is fusion most effective?
 \item \textbf{RQ4:} What are the characteristic failure modes of multimodal clickbait detection in Bengali, and what do they imply for the design of future detectors?
\end{itemize}

Our contributions are four-fold:
\begin{itemize}
 \item \textbf{BanClickThumb dataset.} We present the first publicly available multimodal dataset for Bengali YouTube clickbait detection, comprising 7,147 thumbnail-title pairs collected across five content domains and labeled through a structured ten-annotator pipeline with high inter-annotator agreement (Cohen's Kappa up to 0.93).
 \item \textbf{Comprehensive benchmarks.} We benchmark three multilingual text encoders (mBERT, XLM-RoBERTa, DistilBERT) and three vision transformers (ViT, Swin Transformer, SwiftFormer) in unimodal settings, and all nine vision-text combinations under early, late, and intermediate fusion, 27 multimodal configurations in total.
 \item \textbf{BanClickFusionFormer.} We propose an intermediate fusion architecture pairing ViT with XLM-RoBERTa that achieves the best overall accuracy of 0.84, outperforming the strongest text-only (0.82) and image-only (0.68) baselines.
 \item \textbf{Error analysis and implications.} We conduct a qualitative error analysis that identifies dense visual text, metaphorical language, and culturally specific slang as recurring failure modes, and we distill theoretical and practical implications for information management on low-resource social platforms.
\end{itemize}

The rest of this paper is organized as follows. Section~\ref{sec:background} introduces the preliminary concepts underlying our frameworks. Section~\ref{sec:related} reviews related work and identifies the research gaps. Section~\ref{sec:dataset} describes the construction of BanClickThumb. Section~\ref{sec:methodology} details the proposed methodology. Section~\ref{sec:experiments} presents the experimental setup and results. Section~\ref{sec:discussion} answers the research questions and discusses theoretical and practical implications. Section~\ref{sec:limitations} outlines limitations and future research directions, and Section~\ref{sec:conclusions} concludes the paper.

%%%%%%%%%%%%%%%%%%%%%%%%
%%%%%%%%%%%%%%%%%%%%%%%%
\section{Background}
\label{sec:background}

In this section, we discuss several preliminary concepts to facilitate an understanding of the model families, fusion paradigms, and evaluation criteria used throughout this work. Detecting Bengali clickbait is challenging on both fronts of the multimodal signal: the language exhibits rich morphology and frequent code-mixing with English (``Banglish''), and thumbnails employ sophisticated visual manipulations such as text overlays, exaggerated expressions, and oversaturated color grading.

\subsection{Multilingual Text Encoders}
Transformer-based language models \citep{vaswani2017attention} underpin modern text classification. We consider three multilingual encoders suited to a low-resource language such as Bengali.

\textbf{mBERT.} Multilingual BERT \citep{devlin2018bert} extends BERT to 104 languages, pre-trained with masked language modeling and next-sentence prediction. Its cross-lingual representations allow it to recognize exaggerated or ambiguous Bengali phrasing even with limited task-specific data.

\textbf{XLM-RoBERTa.} XLM-RoBERTa \citep{conneau2019unsupervised} scales cross-lingual pretraining to 2.5\,TB of CommonCrawl text using masked language modeling only. Its stronger cross-lingual generalization makes it well suited to detecting sensationalism and code-mixed cues in Bengali titles.

\textbf{DistilBERT.} DistilBERT \citep{sanh2019distilbert} compresses BERT via knowledge distillation, retaining roughly 97\% of its language understanding while being 40\% smaller and 60\% faster, making it attractive for deployment-constrained moderation pipelines.

\subsection{Vision Transformers}
\textbf{ViT.} The Vision Transformer \citep{dosovitskiy2020image} treats an image as a sequence of patches and applies global self-attention, enabling it to relate distant thumbnail elements, for example, a text overlay in one corner and a shocked facial expression in another, that jointly signal clickbait.

\textbf{Swin Transformer.} The Swin Transformer \citep{liu2021swin} computes self-attention within shifted local windows arranged hierarchically, capturing both fine-grained details (embedded text, facial exaggeration) and global thumbnail composition.

\textbf{SwiftFormer.} SwiftFormer \citep{shaker2023swiftformer} replaces quadratic self-attention with an efficient additive attention mechanism, offering near-real-time inference suitable for on-device or large-scale screening scenarios.

\subsection{Multimodal Fusion Paradigms}
Multimodal learning integrates heterogeneous signals, here, titles and thumbnails, to capture patterns that no single modality fully represents \citep{baltrusaitis2019multimodal}. Fusion can occur at three stages of a network. \emph{Early fusion} concatenates modality features at the input of the joint network, learning cross-modal correlations from the outset. \emph{Late fusion} combines the decisions of independently trained unimodal classifiers, preserving modality specialization. \emph{Intermediate fusion} merges modality representations at hidden layers, allowing the network to model non-linear cross-modal interactions, such as a mismatch between a serious title and a sensational thumbnail, before classification. Section~\ref{sec:methodology} formalizes all three.

\subsection{Evaluation Metrics}
We evaluate all models with four standard classification metrics computed from true positives ($TP$), true negatives ($TN$), false positives ($FP$), and false negatives ($FN$) \citep{goutte2005probabilistic, sokolova2006beyond}:
\begin{equation}
\text{Accuracy} = \frac{TP + TN}{TP + TN + FP + FN}, \qquad
\text{Precision} = \frac{TP}{TP + FP},
\end{equation}
\begin{equation}
\text{Recall} = \frac{TP}{TP + FN}, \qquad
\text{F1} = 2 \cdot \frac{\text{Precision} \cdot \text{Recall}}{\text{Precision} + \text{Recall}}.
\end{equation}
Precision limits the wrongful flagging of legitimate videos, recall captures the fraction of deceptive content caught, and the F1 score balances both, which is particularly informative given the moderate class skew in our data.

%%%%%%%%%%%%%%%%%%%%%%%%
%%%%%%%%%%%%%%%%%%%%%%%%
\section{Related Work}
\label{sec:related}

Clickbait detection has advanced along two methodological tracks: unimodal approaches that rely on a single information source (typically text), and multimodal approaches that integrate complementary signals such as thumbnails, audio, and user behavior. We review both tracks with an emphasis on Bengali and other low-resource settings, and close by synthesizing the research gaps that motivate this work. Tables~\ref{tab:unimodal_datasets} and~\ref{tab:multimodal_datasets} summarize the reviewed studies.

\subsection{Unimodal Clickbait Detection}
Unimodal detection classifies clickbait from a single source, most often the headline or title, because such text carries strong linguistic cues of exaggeration and forward reference \citep{chakraborty2016stop, blom2015click}. Its central weakness is structural: on platforms such as YouTube, where thumbnails coexist with titles, a large portion of the deceptive signal is simply never observed.

The most substantial Bengali resource to date is \textbf{BaitBuster-Bangla} \citep{baitbuster2024}, comprising 253,070 YouTube video instances with titles, descriptions, and thumbnail URLs. The authors fine-tuned several multilingual language models, with XLM-R achieving 99.0\% test accuracy and an F1-macro of 0.990. However, the benchmark exploits only textual features; the thumbnails themselves are provided as URLs and remain unmodeled, leaving fully multimodal detection unexplored. \citet{gothankar2022clickbait} classified YouTube titles using representations ranging from Word2Vec to BERT and DistilBERT embeddings, with DistilBERT + MLP reaching an F1 of 0.92; thumbnails were ignored, and the dataset was not released, restricting reproducibility. For Bengali specifically, \citet{arfat2024bangla} ensembled XLM-RoBERTa, IndicBERT, and BanglaBERT via weighted averaging to reach 96\% accuracy, yet the dataset, its size, and its class balance remain undisclosed, and the approach is again text-only.

Beyond Bengali, \citet{ahmad2020experimental} compared classical models (SVM, logistic regression, na\"ive Bayes) and deep architectures (LSTM, PNN) on roughly 32,000 news articles, with BERT dominating at approximately 97.7\% accuracy. \citet{Potthast2016Corpus} showed on the Webis Clickbait Corpus that stacking ensembles over TF-IDF features marginally outperform single classifiers, yet improvements plateaued without visual signals. Similarly, \citet{alkhassaweneh2023automatic} achieved approximately 98\% accuracy on 32,000 headlines with ensemble classifiers grounded in semantic text analysis, an approach that cannot transfer to settings where the deception resides in imagery.

Taken together, these studies confirm that transformer-based text models achieve high accuracy on headline-style clickbait, yet they systematically neglect the visual channel through which much of modern platform clickbait operates, a limitation that motivates the multimodal benchmarks in this paper.

\begin{table*}[ht]
\centering
\caption{Summary of related works on unimodal clickbait detection.}
\label{tab:unimodal_datasets}
\resizebox{\textwidth}{!}{%
\begin{tabular}{p{0.20\textwidth} p{0.20\textwidth} p{0.25\textwidth} p{0.15\textwidth} p{0.25\textwidth}}
\toprule
\textbf{Reference} & \textbf{Dataset Source} & \textbf{Methodology} & \textbf{Performance} & \textbf{Limitations} \\ \midrule
BaitBuster-Bangla \citep{baitbuster2024} & 253,070 YouTube videos (Bangla) & Fine-tuned XLM-R (multilingual) & Acc: 99.0\% \newline F1: 0.990 & Text-only benchmark; thumbnail visuals unmodeled. \\ \midrule
\citet{gothankar2022clickbait} & YouTube videos (private) & DistilBERT embeddings + MLP & F1: 0.92 & Ignored thumbnails; dataset not publicly available. \\ \midrule
\citet{arfat2024bangla} & Undisclosed (Bangla) & Weighted ensemble (XLM-RoBERTa, IndicBERT, BanglaBERT) & Acc: 96\% & Text-only; dataset size and class balance undisclosed. \\ \midrule
\citet{ahmad2020experimental} & 32,000 news articles & BERT (deep learning) & Acc: $\sim$97.7\% & Limited to textual analysis; no multimodal fusion. \\ \midrule
\citet{Potthast2016Corpus} & Webis Clickbait Corpus & TF-IDF + stacking ensemble & Competitive & Traditional features; gains plateaued without visuals. \\ \midrule
\citet{alkhassaweneh2023automatic} & 32,000 headlines & Ensemble ML classifiers & Acc: $\sim$98\% & Semantic text analysis only; blind to visual cues. \\ \bottomrule
\end{tabular}%
}
\end{table*}

\subsection{Multimodal Clickbait Detection}
Multimodal detection integrates complementary sources, titles, thumbnails, audio, and behavioral signals, into a unified framework \citep{baltrusaitis2019multimodal}, reflecting the reality that platforms like YouTube orchestrate linguistic and visual cues jointly to influence audiences.

\textbf{BaitRadar} \citep{baitradar2021} fused titles, thumbnails, and audio features from YouTube videos through a multi-model deep learning pipeline, achieving 85.34\% accuracy and demonstrating that heterogeneous signals outperform unimodal baselines; its evaluation, however, was confined to English content. \citet{yu2024multimodal} proposed causal representation inference to de-confound dataset biases in title-thumbnail fusion, achieving state-of-the-art results on three English datasets (CLDInst, Clickbait17, FakeNewsNet), yet without addressing low-resource languages. \citet{varshney2021} combined cognitive evidences, video content features from transcripts, human consensus from comments, and user profiling, reaching 93.23\% accuracy on the Fake Video Corpus and 91.48\% on a self-curated Misleading Video Dataset, again exclusively for English.

For Bengali, directly comparable multimodal clickbait work does not yet exist, yet adjacent tasks demonstrate both feasibility and difficulty. \textbf{MultiBanFakeDetect} \citep{faria2025multibanfakedetect} fused textual and visual features for Bangla fake news detection in this journal, confirming the value of fusion for deceptive-content tasks in under-resourced contexts. In multimodal sentiment analysis of Bengali memes, \textbf{MemoSen} \citep{memosen2023} paired ResNet50/DenseNet121 image features with BanglishBERT, while \textbf{SentimentFormer} \citep{sentimentformer2025} fused mBERT/DistilBERT with SwiftFormer via intermediate fusion to reach 79.04\% accuracy, though both report persistent difficulties with label imbalance, dialectal diversity, and figurative language. These sentiment-oriented studies transfer methodological insight yet do not target clickbait, whose deceptive intent requires reasoning about the \emph{mismatch} between promise and content.

\begin{table*}[htbp]
 \centering
 \caption{Summary of related works on multimodal detection.}
 \label{tab:multimodal_datasets}
 \resizebox{\textwidth}{!}{%
\begin{tabular}{p{0.18\linewidth} p{0.15\linewidth} p{0.30\linewidth} p{0.30\linewidth}}
 \toprule
 \textbf{Study} & \textbf{Focus} & \textbf{Methodology \& Performance} & \textbf{Limitations} \\
 \midrule
 BaitRadar \citep{baitradar2021} & Clickbait detection (YouTube) & Title + thumbnail + audio cues; \textbf{85.34\%} accuracy. & English only; fusion unexplored for Bangla. \\
 \midrule
 \citet{yu2024multimodal} & Clickbait detection & Causal representation inference to remove dataset bias; SOTA on Clickbait17 \& FakeNewsNet. & English datasets; low-resource languages unaddressed. \\
 \midrule
 \citet{varshney2021} & Clickbait video detection & Cognitive evidence framework (content, comments, user profile); \textbf{93.23\%} (FVC), \textbf{91.48\%} (MVD). & Confined to English; no Bangla multimodal dataset. \\
 \midrule
 MultiBanFakeDetect \citep{faria2025multibanfakedetect} & Fake news detection (Bangla) & Advanced text-image fusion for multimodal Bangla fake news. & Targets fake news, not thumbnail-title clickbait. \\
 \midrule
 MemoSen \citep{memosen2023} & Sentiment analysis (Bangla memes) & ResNet50/DenseNet121 + BanglishBERT. & Sentiment task; no deceptive-intent labels. \\
 \midrule
 SentimentFormer \citep{sentimentformer2025} & Sentiment analysis (Bangla memes) & mBERT/DistilBERT + SwiftFormer, intermediate fusion; \textbf{79.04\%} accuracy. & Sentiment task; imbalance and sarcasm challenges. \\
 \bottomrule
 \end{tabular}
}
\end{table*}

\subsection{Research Gaps and Our Positioning}
Several research gaps emerge from this synthesis: (i) no publicly available, manually annotated multimodal dataset exists for Bengali YouTube clickbait detection; (ii) existing Bengali clickbait detectors are exclusively text-based, leaving the visual channel of deception unmodeled; (iii) no systematic comparison of early, late, and intermediate fusion has been conducted for clickbait detection in any low-resource language; and (iv) the failure modes of multimodal detectors on culturally specific, code-mixed content remain uncharacterized. Our proposed BanClickThumb dataset and the BanClickTextFormer, BanClickImageFormer, and BanClickFusionFormer benchmarks address all of these gaps.

%%%%%%%%%%%%%%%%%%%%%%%%
%%%%%%%%%%%%%%%%%%%%%%%%
\section{The BanClickThumb Dataset}
\label{sec:dataset}

\begin{figure*}[htbp]
 \centering
 \includegraphics[width=\linewidth]{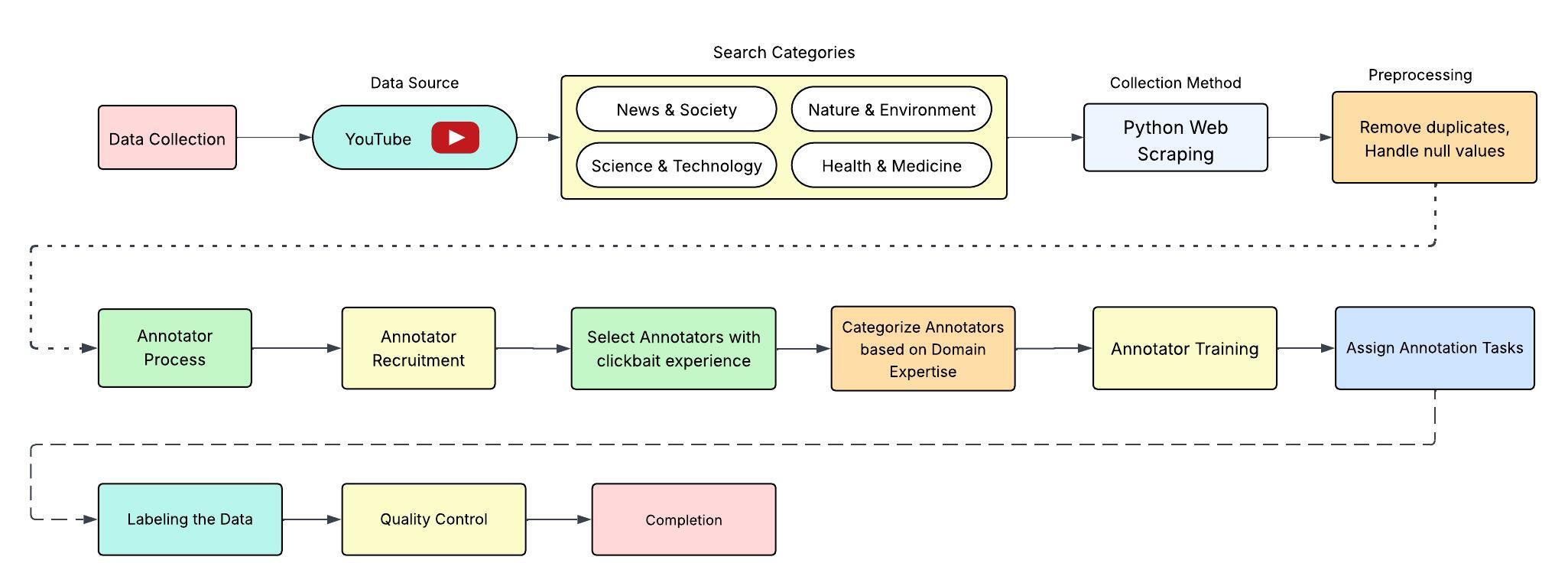}
 \caption{Step-by-step process for BanClickThumb dataset creation: data collection, preprocessing, annotation, quality control, and validation.}
 \label{fig:dataset_pipeline}
\end{figure*}

This section answers the construction side of RQ1 by describing how BanClickThumb was collected, annotated, validated, and partitioned, and how it compares with existing clickbait resources. Figure~\ref{fig:dataset_pipeline} provides an overview of the full pipeline.

\subsection{Data Collection}
YouTube was chosen as the source platform owing to its dominant position in Bengali digital media consumption \citep{youtube_bd_users}. Thumbnails and titles of publicly available videos were collected through automated web scraping \citep{Mitchell2018}, in accordance with YouTube's fair-use provisions and with care taken to avoid collecting personally identifiable information (PII).

\begin{figure*}[!t]
 \centering
 \includegraphics[width=1\textwidth]{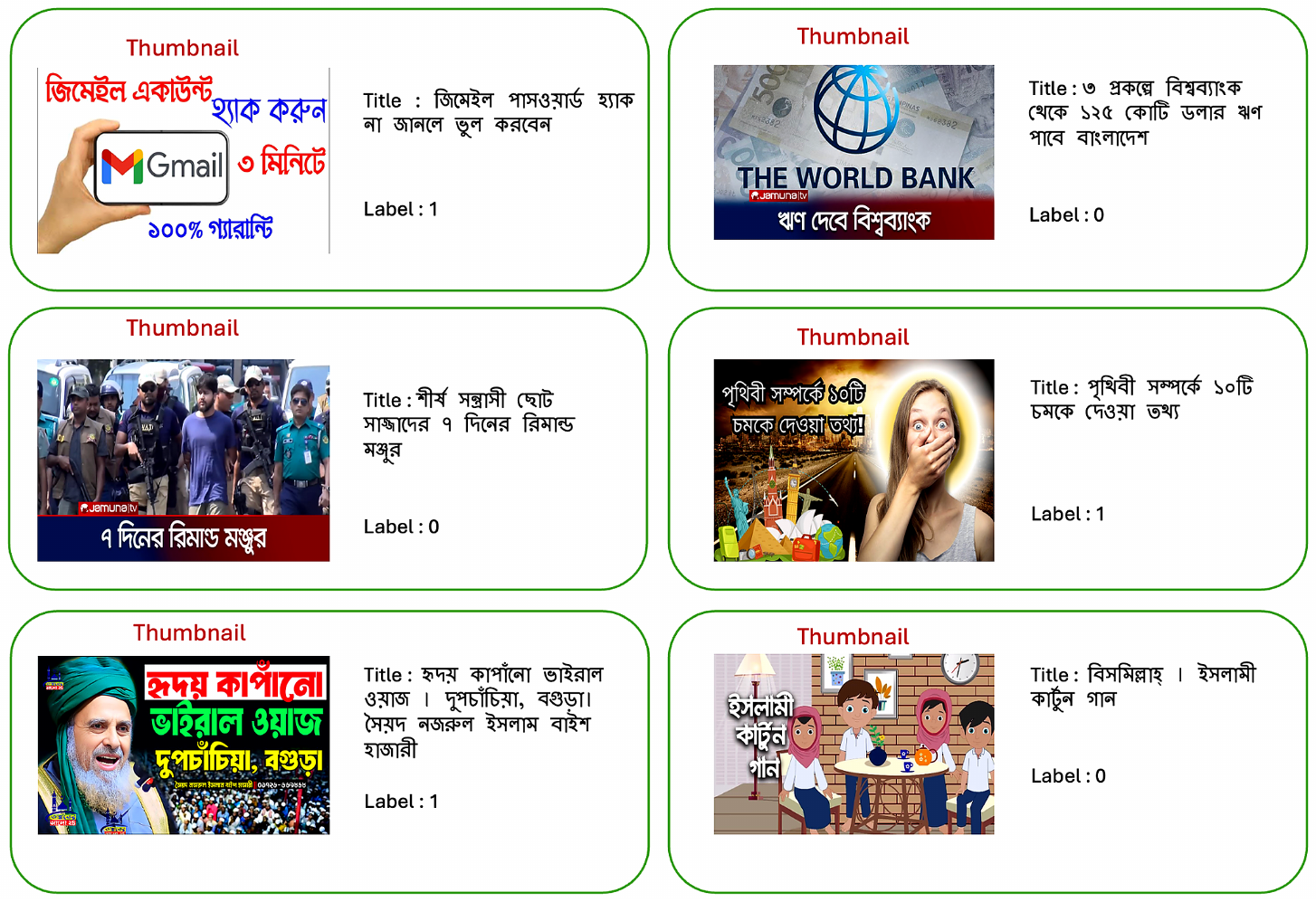}
 \caption{Representative thumbnail-title pair examples with binary labels (0 = Non-clickbait, 1 = Clickbait).}
 \label{fig:dataset_examples}
\end{figure*}

To capture the semantic diversity and visual variation typical of real-world clickbait, videos were sampled from five content domains: \textbf{News and Politics} (political leaders, crises, wars, crime), \textbf{Health and Lifestyle} (medical tips, pandemics, food, fitness), \textbf{Science and Technology} (machinery, space technology, gadgets), \textbf{General Knowledge} (geography, agriculture, history), and \textbf{Entertainment}. Samples were labeled with a binary scheme, clickbait (1) and non-clickbait (0), and approximate class balance was maintained during collection to mitigate learning bias \citep{Buda2018}. Figure~\ref{fig:dataset_examples} shows representative examples from both classes.

\subsection{Data Annotation}
The annotation protocol was designed to maximize reliability and consistency, and proceeded in five steps.

\textbf{(i) Annotator recruitment.} Ten annotators were recruited based on Bangla proficiency, prior annotation experience, and domain knowledge in news, politics, technology, and entertainment. Table~\ref{tab:annotator_profiles} summarizes their profiles.

\begin{table*}[ht]
\centering
\caption{Summary of annotator profiles involved in the study.}
\label{tab:annotator_profiles}
\resizebox{\textwidth}{!}{%
\begin{tabular}{p{0.20\textwidth} p{0.25\textwidth} p{0.20\textwidth} p{0.30\textwidth}}
\toprule
\textbf{Annotator ID} & \textbf{Qualifications} & \textbf{Experience} & \textbf{Expertise} \\ \midrule
Annotator 1 & Undergraduate & 1 year & Multimodal clickbait \\
Annotator 2 & Undergraduate & 1 year & Multimodal clickbait \\
Annotator 3 & Undergraduate & 2 years & News analysis \\
Annotator 4 & Graduate & 1 year & Social media content \\
Annotator 5 & Graduate & 3 years & Visual content analysis \\
Annotator 6 & Undergraduate & 2 years & Clickbait identification \\
Annotator 7 & Undergraduate & 2 years & Multimodal research \\
Annotator 8 & Undergraduate & 1 year & News and entertainment \\
Annotator 9 & Graduate & 2 years & Technology content \\
Annotator 10 & Graduate & 2 years & Multimodal content \\ \bottomrule
\end{tabular}%
}
\end{table*}

\textbf{(ii) Training and guidelines.} Annotators were grouped by domain expertise and trained with written guidelines defining clickbait as content whose title or thumbnail is exaggerated, misleading, or unfulfilled by the video, and non-clickbait as factually representative content \citep{Chen2015}. Worked examples clarified boundary cases.

\textbf{(iii) Annotation.} Each thumbnail-title pair was independently reviewed by at least three annotators from diverse backgrounds, ensuring that no single perspective dominated the final label.

\textbf{(iv) Quality control.} Annotation was monitored continuously and disagreements were resolved through structured discussion. Inter-annotator agreement, measured with Cohen's Kappa and Fleiss' Kappa \citep{Cohen1960, Fleiss1971}, ranged from 0.83 to 0.93, indicating near-perfect reliability by conventional standards \citep{Artstein2008}.

\textbf{(v) Validation.} Final labels (0 = Non-clickbait, 1 = Clickbait) were assigned by majority consensus, following established practice for mitigating annotation noise \citep{Beigman2014}.

\subsection{Dataset Splitting}
\begin{figure}[htbp]
\centering
\includegraphics[width=\columnwidth]{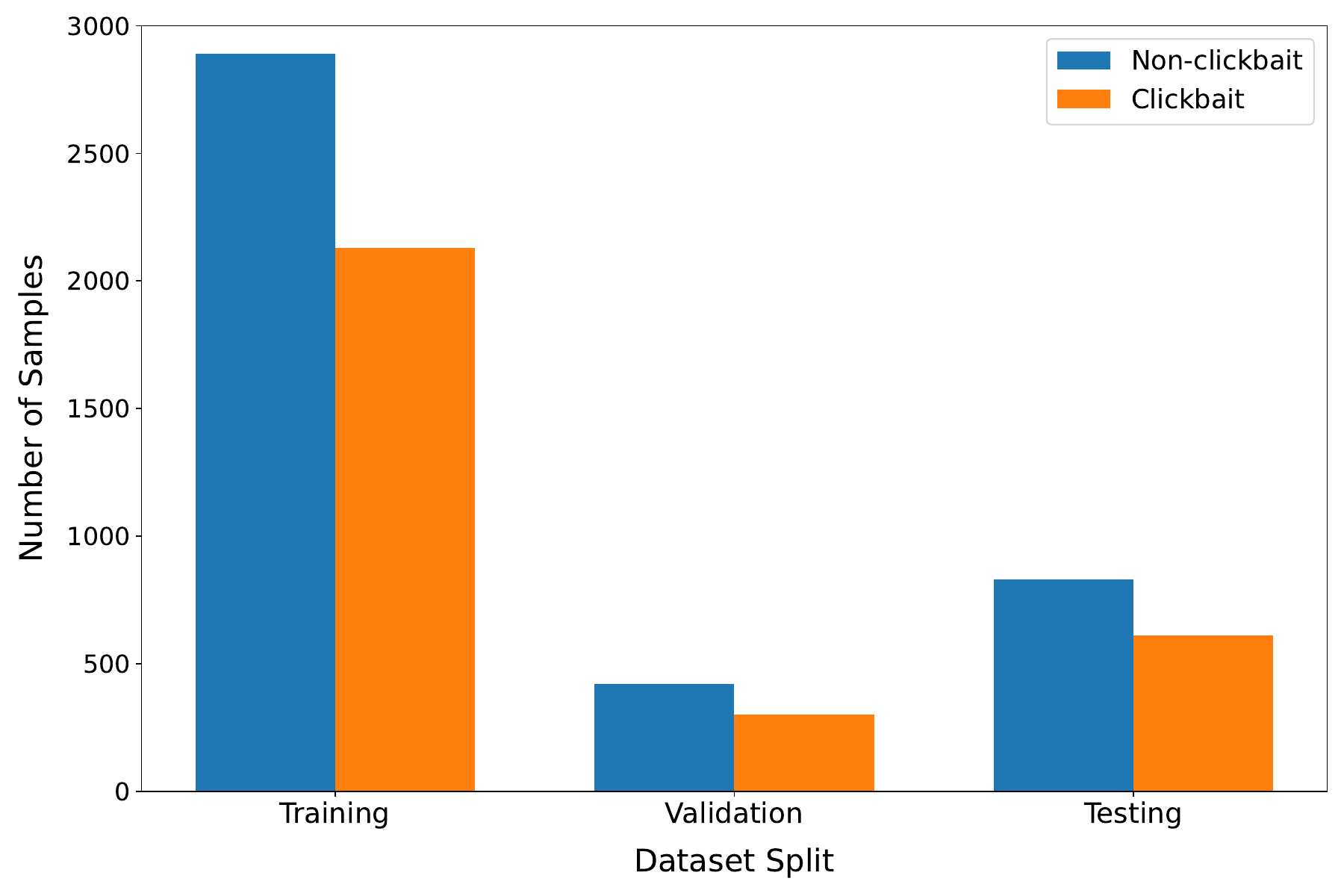}
\caption{Label-wise distribution across training, validation, and test sets.}
\label{fig:label_distribution}
\end{figure}

The 7,147 annotated pairs were partitioned with stratified sampling to preserve class proportions \citep{Kohavi1995} into a 70\%/10\%/20\% train/validation/test split: 5,002 training pairs (2,117 clickbait, 2,885 non-clickbait), 715 validation pairs (302 clickbait, 413 non-clickbait), and 1,430 test pairs (605 clickbait, 825 non-clickbait). During model development, five-fold cross-validation on the training partition was additionally used to verify the stability of hyperparameter choices \citep{Zhang2017}. Table~\ref{tab:dataset_split} and Figure~\ref{fig:label_distribution} report the distribution.

\begin{table}[ht]
\centering
\caption{Distribution of thumbnail-title pairs across dataset splits.}
\label{tab:dataset_split}
\begin{tabular}{lccc}
\toprule
\textbf{Set} & \textbf{Non-clickbait} & \textbf{Clickbait} & \textbf{Total} \\
\midrule
Training & 2,885 & 2,117 & 5,002 \\
Validation & 413 & 302 & 715 \\
Testing & 825 & 605 & 1,430 \\
\midrule
\textbf{Total} & \textbf{4,123} & \textbf{3,024} & \textbf{7,147} \\
\bottomrule
\end{tabular}
\end{table}

\subsection{Comparison with Existing Datasets}
Table~\ref{tab:comparative_simple} positions BanClickThumb against existing clickbait resources. BaitBuster-Bangla \citep{baitbuster2024} is far larger (253,070 samples) yet text-only, providing thumbnail URLs rather than annotated images, and therefore cannot support direct multimodal fusion. The BaitRadar dataset \citep{baitradar2021} covers text, image, and audio yet contains only English content, and the Webis Clickbait Corpus 2016 \citep{Potthast2016Corpus} is a small, imbalanced, text-only English resource. BanClickThumb is, to our knowledge, the first large-scale, approximately balanced, manually annotated multimodal dataset for Bengali clickbait detection.

\begin{table*}[ht]
\centering
\caption{Comparative analysis of BanClickThumb with existing datasets.}
\label{tab:comparative_simple}
\begin{tabular}{l l l l l}
\toprule
\textbf{Dataset} & \textbf{Modality} & \textbf{\#Samples} & \textbf{Label Type} & \textbf{Language} \\
\midrule
\textbf{BanClickThumb (Ours)} & Text + Image & \textbf{7,147} & Binary (0/1) & Bangla \\
BaitBuster-Bangla & Text only & 253,070 & Binary (0/1) & Bangla \\
BaitRadar & Text + Image + Audio & 14,000 & Binary (0/1) & English \\
Webis Corpus 16 & Text only & 2,992 & Binary (0/1) & English \\
\bottomrule
\end{tabular}
\end{table*}

%%%%%%%%%%%%%%%%%%%%%%%%
%%%%%%%%%%%%%%%%%%%%%%%%
\section{Methodology}
\label{sec:methodology}

\begin{figure*}[htpb]
 \centering
 \includegraphics[width=\textwidth]{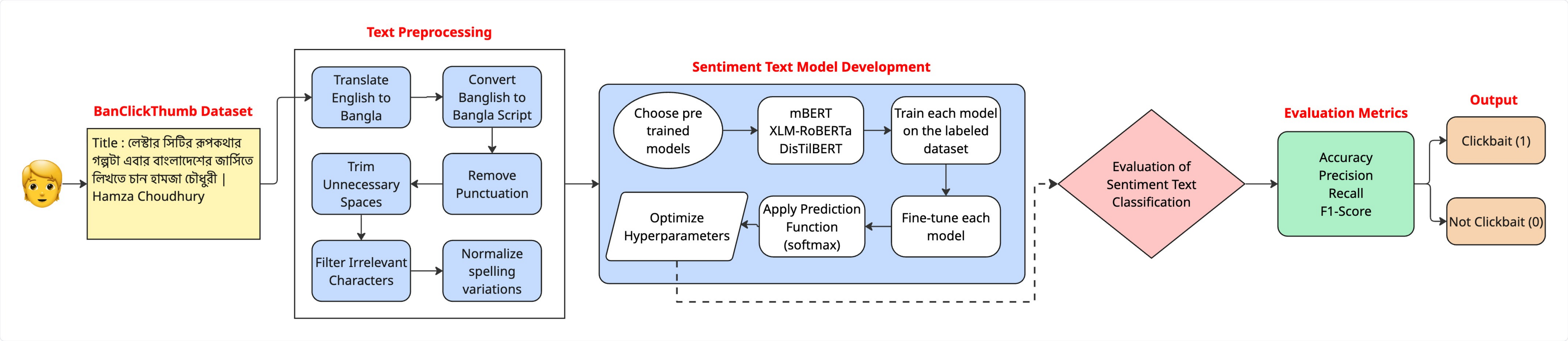}
 \caption{Unimodal clickbait classification framework for Bangla YouTube video titles (BanClickTextFormer).}
 \label{fig:text_framework}
\end{figure*}

\begin{figure*}[htpb]
 \centering
 \includegraphics[width=\textwidth]{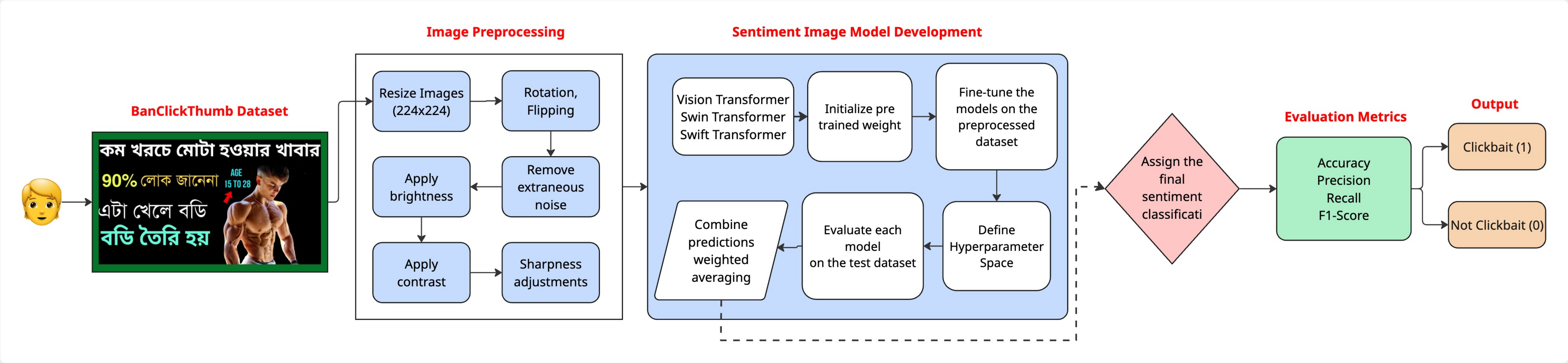}
 \caption{Unimodal clickbait classification framework for YouTube video thumbnails (BanClickImageFormer).}
 \label{fig:image_framework}
\end{figure*}

\begin{figure*}[htpb]
 \centering
 \includegraphics[width=\textwidth]{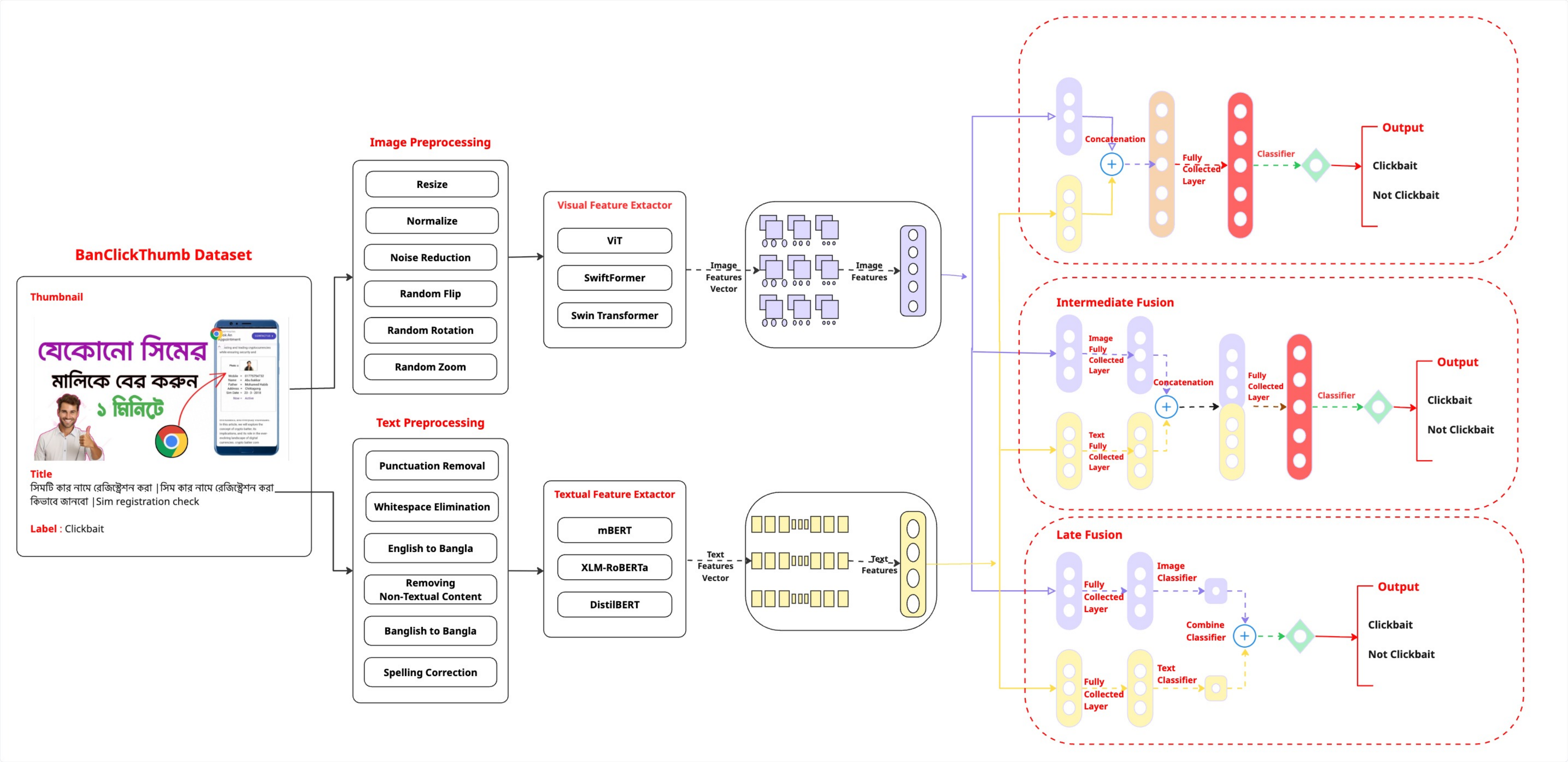}
 \caption{Multimodal fusion framework for clickbait detection using YouTube titles and thumbnails (BanClickFusionFormer).}
 \label{fig:fusion_framework}
\end{figure*}

In this section, we discuss the technical details of our three proposed frameworks. Let the dataset be $\mathcal{D} = \{(t_i, v_i, y_i)\}_{i=1}^{N}$, where $t_i$ is a Bengali title, $v_i$ the corresponding thumbnail, and $y_i \in \{0, 1\}$ the clickbait label. \textbf{BanClickTextFormer} classifies from $t_i$ alone, \textbf{BanClickImageFormer} from $v_i$ alone, and \textbf{BanClickFusionFormer} from the pair $(t_i, v_i)$ through early, late, and intermediate fusion. Figures~\ref{fig:text_framework} and~\ref{fig:image_framework} illustrate the unimodal architectures, and Figure~\ref{fig:fusion_framework} presents the fusion framework. All code, including the fusion implementations, dimension-alignment steps, and hyperparameter schedules, is publicly available in our GitHub repository\footnote{\url{https://github.com/arifbaust10/BanClickThumb-A-Multimodal-Dataset-and-Benchmarks-for-Clickbait-Detection-in-Bengali-YouTube-Videos}} to ensure transparency and reproducibility.

\subsection{BanClickTextFormer: Unimodal Framework for Bengali Titles}

\subsubsection{Text Preprocessing}
Bengali YouTube titles pose distinct challenges: rich morphology, flexible syntax, pervasive code-mixing, and deliberately sensational punctuation. We apply a systematic pipeline to each title $d_i$. First, embedded English tokens are translated into Bangla using a translation function $T(\cdot)$, and Romanized Bangla (``Banglish'') is transliterated into standard Bangla script, ensuring script uniformity; we acknowledge that translation may not perfectly preserve English slang engineered to drive clicks. Second, punctuation is normalized: exaggerated marks (e.g., repeated exclamation or question marks) are reduced unless they form patterns indicative of curiosity gaps, and extraneous whitespace is removed. Third, dialectal and phonetic spelling variants are normalized to ensure that semantically identical words share a representation, and irrelevant symbols and control characters are filtered. Formally,
\begin{equation}
\tilde{d}_i = \mathrm{FilterChars}\big(\mathrm{TrimSpaces}(\mathrm{NormPunct}(T(d_i)))\big),
\end{equation}
yielding the cleaned corpus $\mathcal{D}' = \{\tilde{d}_1, \dots, \tilde{d}_N\}$.

\subsubsection{Model Development}
We fine-tune \textbf{XLM-RoBERTa} on the 5,002 cleaned training titles for binary classification. Given the pre-trained model $M$ and labeled data $L$, fine-tuning produces $M^{*} = F(M, L)$, and predictions follow
\begin{equation}
P(y \mid \tilde{d}_i) = \mathrm{softmax}\big(M^{*}(\tilde{d}_i)\big).
\end{equation}
Because the training set is moderately skewed (2,117 clickbait vs.\ 2,885 non-clickbait), class-weighted cross-entropy is used to ensure that the model attends adequately to the clickbait class. Hyperparameters, learning rate $\eta$, batch size $B$, dropout $\lambda$, and epochs $E$, are selected by grid search over the validation set:
\begin{equation}
h^{*} = \operatorname*{arg\,max}_{h \in \mathcal{H}} \; \mathrm{Evaluate}(M^{*}, h, \mathcal{D}_{val}).
\end{equation}

\subsection{BanClickImageFormer: Unimodal Framework for Thumbnails}

\subsubsection{Image Preprocessing}
Thumbnails contain complex attention-engineering devices: superimposed text, exaggerated facial expressions, and high-contrast graphics. Every thumbnail $I_i$ is resized to $224 \times 224$ pixels for compatibility with transformer backbones, augmented with random rotation, horizontal flipping, and brightness adjustment to improve generalization, and enhanced by contrast/sharpness normalization to ensure that the model focuses on semantic content rather than compression artifacts:
\begin{equation}
I'_i = \mathrm{Enhance}\big(\mathrm{Augment}(\mathrm{Resize}(I_i, 224 \times 224))\big).
\end{equation}

\subsubsection{Model Development}
We fine-tune \textbf{SwiftFormer} on the 5,002 training thumbnails, selected among the vision backbones for its favorable speed-accuracy balance (Section~\ref{sec:experiments} reports all three backbones). During fine-tuning, the model learns visual clickbait tropes, shock reactions, misleading arrows, suggestive imagery, against the more neutral composition of non-clickbait thumbnails. The validation set (715 images) monitors training, and prediction follows
\begin{equation}
\hat{y}_i = \operatorname*{arg\,max}_{k \in \{0,1\}} P(y = k \mid I'_i).
\end{equation}
Hyperparameters are tuned by grid search as in the text branch.

\subsection{BanClickFusionFormer: Multimodal Fusion Strategies}
\label{sec:fusion}

\subsubsection{Feature Extraction}
Both modalities are first encoded independently. The text encoder $f_{text}(\cdot)$ (XLM-RoBERTa) produces contextual representations that capture the semantics of sensational, code-mixed Bengali phrasing, and the vision encoder $f_{image}(\cdot)$ (ViT) produces patch-level representations that capture both global thumbnail composition and local manipulations. Let $\phi_{text}(\cdot)$ and $\phi_{image}(\cdot)$ denote the corresponding projection/activation functions applied to align dimensions. Rich features from both encoders allow the fusion stages to detect \emph{discrepancies between what the title promises and what the thumbnail depicts}, the core signature of clickbait.

\subsubsection{Fusion Techniques}
We systematically compare three fusion strategies.

\paragraph{(a) Early Fusion.} Feature vectors from both encoders are concatenated before any joint classification layers, creating a shared representation space from the outset:
\begin{equation}
z_{early} = f_{fusion}\big(\big[\phi_{text}(f_{text}(X));\; \phi_{image}(f_{image}(Y))\big]\big),
\label{eq:early}
\end{equation}
where $[\cdot\,;\cdot]$ denotes concatenation and $z_{early}$ feeds the classifier.

\paragraph{(b) Late Fusion.} The independently trained unimodal models produce class probabilities that are combined at the decision level:
\begin{equation}
P_{fusion} = \alpha \cdot P_{text}\big(f_{text}(X)\big) + (1 - \alpha) \cdot P_{image}\big(f_{image}(Y)\big),
\label{eq:late}
\end{equation}
where the weight $\alpha \in [0,1]$ balances the modalities and is tuned on the validation set, revealing which modality carries more predictive power.

\paragraph{(c) Intermediate Fusion (BanClickFusionFormer).} Intermediate representations from both pipelines are merged at hidden layers, allowing the network to learn non-linear cross-modal interactions before classification:
\begin{equation}
Z_{fusion} = f_{fusion}\big(\phi_{text}(f_{text}(X)),\; \phi_{image}(f_{image}(Y))\big).
\label{eq:intermediate}
\end{equation}
This is the configuration underlying our proposed \textbf{BanClickFusionFormer}, which pairs ViT (image) with XLM-RoBERTa (text) and, as Section~\ref{sec:experiments} shows, achieves the strongest overall performance.

\subsubsection{Training Objective and Hyperparameter Optimization}
All fusion models minimize the (class-weighted) cross-entropy loss over the training set,
\begin{equation}
\mathcal{L} = \sum_{i=1}^{N} \ell\big(y_i, \hat{y}_i;\, \eta, B, \lambda, \alpha\big),
\end{equation}
with the learning rate, batch size, dropout, and (for late fusion) the weight $\alpha$ tuned to maximize validation accuracy. This tuning prevents over-reliance on a single modality and preserves generalization across the diverse domains in BanClickThumb.

%%%%%%%%%%%%%%%%%%%%%%%%
%%%%%%%%%%%%%%%%%%%%%%%%
\section{Experiments and Result Analysis}
\label{sec:experiments}

In this section, we describe the experimental setup, report the performance of all unimodal and multimodal configurations, and analyze representative failure cases.

\subsection{Experimental Setup}
All models were implemented in Python with the PyTorch framework, and experiments were conducted across three computing environments, Jupyter Notebook (v6.5.5), Kaggle (v1.6.17), and Google Colaboratory, to assess the robustness of the pipeline under different runtimes. The Jupyter and Kaggle environments used Python~3.10 with PyTorch~2.9, while Colaboratory used Python~3.12 with a PyTorch~2.8-series build. The consistency of results across these environments supports the stability of the reported findings.

\subsection{Hyperparameter Settings}
Table~\ref{tab:hyper_image_text} summarizes the settings for the unimodal backbones. The vision models (ViT, Swin Transformer, SwiftFormer) were fine-tuned with a batch size of 32, a learning rate of $2 \times 10^{-4}$, and the AdamW optimizer for 15 epochs; the text models (mBERT, XLM-RoBERTa, DistilBERT) used the same batch size and optimizer with a learning rate of $2 \times 10^{-5}$ for 15 epochs. All 27 fusion configurations, each of the three vision backbones paired with each of the three text encoders under early, late, and intermediate fusion, were trained with a uniform batch size of 32, a learning rate of $2 \times 10^{-5}$, and the AdamW optimizer for 15 epochs (Table~\ref{tab:fusion_hyper}), ensuring that performance differences are attributable to architecture rather than tuning disparities.

\begin{table*}[htbp]
\centering
\caption{Hyperparameter settings for the unimodal text and image backbones.}
\label{tab:hyper_image_text}
\begin{tabular}{llcccl}
\toprule
\textbf{Approach} & \textbf{Model} & \textbf{Batch Size} & \textbf{Epochs} & \textbf{LR} & \textbf{Optimizer} \\
\midrule
\multirow{3}{*}{Image-Based}
 & ViT & 32 & 15 & $2 \times 10^{-4}$ & AdamW \\
 & Swin Transformer & 32 & 15 & $2 \times 10^{-4}$ & AdamW \\
 & SwiftFormer & 32 & 15 & $2 \times 10^{-4}$ & AdamW \\
\midrule
\multirow{3}{*}{Text-Based}
 & mBERT & 32 & 15 & $2 \times 10^{-5}$ & AdamW \\
 & XLM-RoBERTa & 32 & 15 & $2 \times 10^{-5}$ & AdamW \\
 & DistilBERT & 32 & 15 & $2 \times 10^{-5}$ & AdamW \\
\bottomrule
\end{tabular}
\end{table*}

\begin{table}[htbp]
\centering
\caption{Hyperparameter settings shared by all early, late, and intermediate fusion configurations (each of ViT, Swin Transformer, and SwiftFormer paired with each of mBERT, XLM-RoBERTa, and DistilBERT).}
\label{tab:fusion_hyper}
\begin{tabular}{lcccc}
\toprule
\textbf{Fusion} & \textbf{Batch Size} & \textbf{Epochs} & \textbf{LR} & \textbf{Optimizer} \\
\midrule
Early & 32 & 15 & $2 \times 10^{-5}$ & AdamW \\
Late& 32 & 15 & $2 \times 10^{-5}$ & AdamW \\
Intermediate & 32 & 15 & $2 \times 10^{-5}$ & AdamW \\
\bottomrule
\end{tabular}
\end{table}

\subsection{Unimodal Performance (RQ2)}
Table~\ref{tab:unimodal_metrics_adjusted} summarizes the unimodal results. Among text models, \textbf{XLM-RoBERTa (BanClickTextFormer)} achieves the highest accuracy of 0.82 (precision 0.81, recall 0.76, F1 0.79), with mBERT and DistilBERT both at 0.78 accuracy. Among vision models, \textbf{SwiftFormer (BanClickImageFormer)} leads with 0.68 accuracy (precision 0.67, recall 0.66, F1 0.67), followed by Swin Transformer (0.67) and ViT (0.64). The 14-point gap between the best text and best image models indicates that titles carry the stronger deceptive signal, though thumbnails clearly contribute meaningful information, a division of labor that motivates fusion.

\begin{table*}[htbp]
\centering
\caption{Performance metrics for unimodal text and image models on the BanClickThumb test set.}
\label{tab:unimodal_metrics_adjusted}
\begin{tabular}{l l c c c c}
\toprule
\textbf{Approach} & \textbf{Model} & \textbf{Accuracy} & \textbf{Precision} & \textbf{Recall} & \textbf{F1} \\
\midrule
Text-based & mBERT & 0.78 & 0.72 & 0.78 & 0.75 \\
Text-based & XLM-RoBERTa (BanClickTextFormer) & \textbf{0.82} & 0.81 & 0.76 & 0.79 \\
Text-based & DistilBERT & 0.78 & 0.75 & 0.72 & 0.74 \\
Image-based & ViT & 0.64 & 0.63 & 0.61 & 0.61 \\
Image-based & Swin Transformer & 0.67 & 0.67 & 0.66 & 0.66 \\
Image-based & SwiftFormer (BanClickImageFormer) & \textbf{0.68} & 0.67 & 0.66 & 0.67 \\
\bottomrule
\end{tabular}
\end{table*}

\subsection{Early Fusion Performance (RQ3)}
Table~\ref{tab:early_fusion_metrics} reports the early fusion results. \textbf{Swin Transformer + XLM-RoBERTa} attains the strongest early fusion performance (accuracy 0.83, weighted F1 0.83), closely followed by SwiftFormer + XLM-RoBERTa (0.83 accuracy, 0.82 F1) and ViT + XLM-RoBERTa (0.82 accuracy, 0.82 F1). Combinations with mBERT cluster at 0.79 to 0.80 accuracy, and DistilBERT pairings trail slightly, with ViT + DistilBERT lowest at 0.78. Across the board, pairing any visual backbone with XLM-RoBERTa yields the most robust early fusion results, echoing the unimodal superiority of XLM-RoBERTa.

\begin{table*}[htbp]
\centering
\caption{Performance metrics for early fusion multimodal models on BanClickThumb.}
\label{tab:early_fusion_metrics}
\begin{tabular}{l c c c c}
\toprule
\textbf{Model} & \textbf{Accuracy} & \textbf{Precision} & \textbf{Recall} & \textbf{Weighted F1} \\
\midrule
Swin Transformer + mBERT & 0.80 & 0.80 & 0.80 & 0.80 \\
Swin Transformer + XLM-RoBERTa & \textbf{0.83} & 0.83 & 0.83 & \textbf{0.83} \\
Swin Transformer + DistilBERT & 0.79 & 0.79 & 0.79 & 0.79 \\
SwiftFormer + mBERT & 0.79 & 0.79 & 0.80 & 0.79 \\
SwiftFormer + XLM-RoBERTa & 0.83 & 0.82 & 0.82 & 0.82 \\
SwiftFormer + DistilBERT & 0.79 & 0.78 & 0.79 & 0.79 \\
ViT + mBERT & 0.80 & 0.80 & 0.80 & 0.80 \\
ViT + XLM-RoBERTa & 0.82 & 0.82 & 0.83 & 0.82 \\
ViT + DistilBERT & 0.78 & 0.81 & 0.76 & 0.77 \\
\bottomrule
\end{tabular}
\end{table*}

\subsection{Late Fusion Performance (RQ3)}
Table~\ref{tab:late_fusion_metrics} summarizes the late fusion results. The three XLM-RoBERTa pairings again lead, each reaching 0.83 accuracy, with Swin Transformer + XLM-RoBERTa achieving the highest weighted F1 (0.87). The mBERT pairings are competitive on accuracy (0.80 to 0.82) yet exhibit noticeably lower recall in some configurations (e.g., ViT + mBERT at 0.65), reflecting a known weakness of decision-level averaging: when the weaker image model disagrees, correct text-model detections can be diluted. DistilBERT pairings remain moderate (0.79 to 0.80 accuracy).

\begin{table*}[htbp]
\centering
\caption{Performance metrics for late fusion multimodal models on BanClickThumb.}
\label{tab:late_fusion_metrics}
\begin{tabular}{l c c c c}
\toprule
\textbf{Model} & \textbf{Accuracy} & \textbf{Precision} & \textbf{Recall} & \textbf{Weighted F1} \\
\midrule
Swin Transformer + mBERT & 0.80 & 0.73 & 0.82 & 0.78 \\
Swin Transformer + XLM-RoBERTa & \textbf{0.83} & 0.88 & 0.89 & \textbf{0.87} \\
Swin Transformer + DistilBERT & 0.80 & 0.80 & 0.71 & 0.75 \\
SwiftFormer + mBERT & 0.82 & 0.84 & 0.72 & 0.77 \\
SwiftFormer + XLM-RoBERTa & 0.83 & 0.80 & 0.80 & 0.80 \\
SwiftFormer + DistilBERT & 0.79 & 0.72 & 0.82 & 0.76 \\
ViT + mBERT & 0.81 & 0.86 & 0.65 & 0.74 \\
ViT + XLM-RoBERTa & 0.83 & 0.83 & 0.76 & 0.79 \\
ViT + DistilBERT & 0.80 & 0.78 & 0.75 & 0.76 \\
\bottomrule
\end{tabular}
\end{table*}

\subsection{Intermediate Fusion Performance (RQ3)}
Table~\ref{tab:intermediate_fusion_metrics} presents the intermediate fusion results. \textbf{ViT + XLM-RoBERTa (BanClickFusionFormer)} achieves the best overall performance in the entire study: 0.84 accuracy with balanced precision and recall of 0.81 (weighted F1 0.81). SwiftFormer + XLM-RoBERTa follows at 0.83 accuracy (F1 0.80). Interestingly, the intermediate Swin Transformer + XLM-RoBERTa configuration collapses to 0.71 accuracy with a recall of only 0.51, despite the same pairing leading the early and late fusion settings, suggesting that Swin's hierarchical window representations align poorly with XLM-RoBERTa's token space at intermediate depths, causing the joint layers to underuse the text signal. This sensitivity of intermediate fusion to representation compatibility is itself an informative finding.

\begin{table*}[htbp]
\centering
\caption{Performance metrics for intermediate fusion multimodal models on BanClickThumb.}
\label{tab:intermediate_fusion_metrics}
\begin{tabular}{l c c c c}
\toprule
\textbf{Model} & \textbf{Accuracy} & \textbf{Precision} & \textbf{Recall} & \textbf{Weighted F1} \\
\midrule
Swin Transformer + mBERT & 0.80 & 0.79 & 0.72 & 0.75 \\
Swin Transformer + XLM-RoBERTa & 0.71 & 0.72 & 0.51 & 0.60 \\
Swin Transformer + DistilBERT & 0.80 & 0.80 & 0.71 & 0.75 \\
SwiftFormer + mBERT & 0.81 & 0.81 & 0.73 & 0.77 \\
SwiftFormer + XLM-RoBERTa & 0.83 & 0.78 & 0.81 & 0.80 \\
SwiftFormer + DistilBERT & 0.79 & 0.76 & 0.76 & 0.76 \\
ViT + mBERT & 0.82 & 0.82 & 0.73 & 0.77 \\
ViT + XLM-RoBERTa (BanClickFusionFormer) & \textbf{0.84} & 0.81 & 0.81 & \textbf{0.81} \\
ViT + DistilBERT & 0.79 & 0.75 & 0.77 & 0.76 \\
\bottomrule
\end{tabular}
\end{table*}

\begin{figure*}[htbp]
 \centering
 \includegraphics[width=\linewidth]{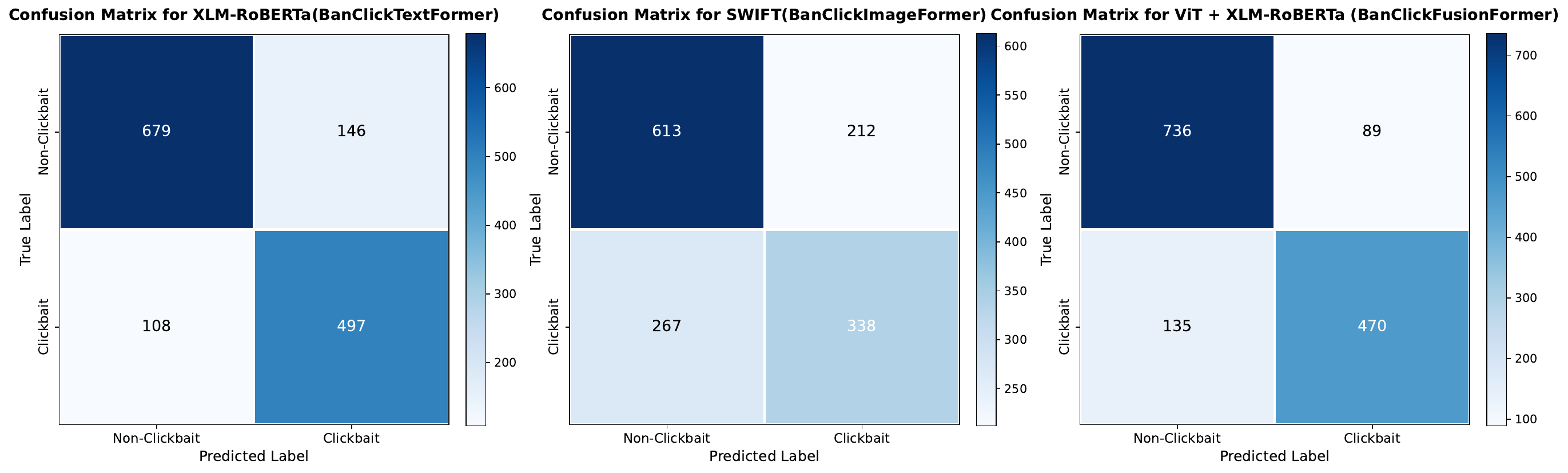}
 \caption{Confusion matrices of BanClickTextFormer, BanClickImageFormer, and BanClickFusionFormer on the BanClickThumb test set.}
 \label{fig:confusion_matrices}
\end{figure*}

Figure~\ref{fig:confusion_matrices} shows the confusion matrices of the best model in each setting. BanClickTextFormer misclassifies relatively few instances, BanClickImageFormer performs moderately, and BanClickFusionFormer exhibits the most balanced classification across both classes, confirming the benefit of combining textual and visual features. Overall, BanClickFusionFormer improves upon the text-only baseline by 2 percentage points and upon the image-only baseline by 16 percentage points.

\subsection{Error Analysis (RQ4)}
\label{sec:error}
\begin{figure}[htbp]
 \centering
 \includegraphics[width=\linewidth]{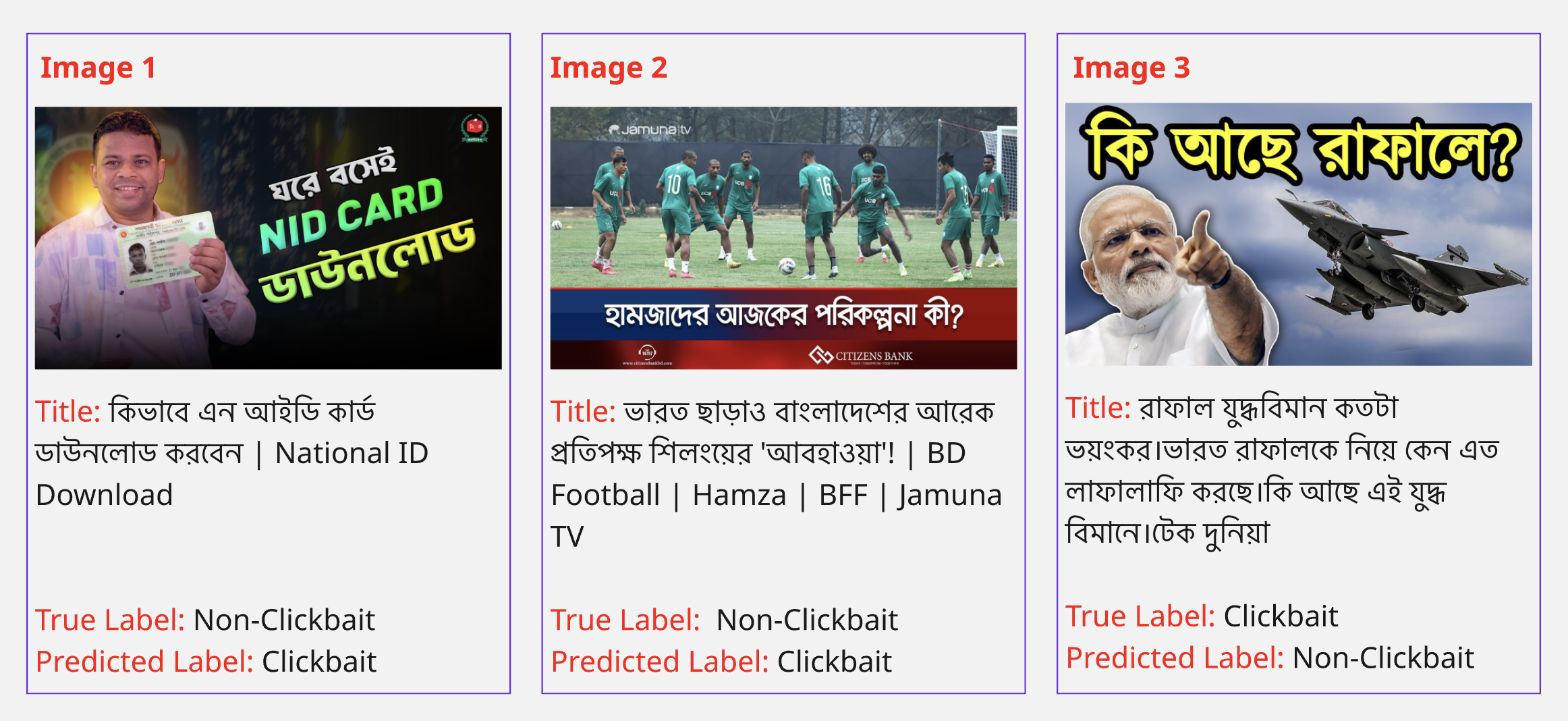}
 \caption{Representative error cases of multimodal clickbait classification in Bengali YouTube videos.}
 \label{fig:error_analysis}
\end{figure}

To understand where multimodal detection fails, we qualitatively examined misclassified test instances. Three recurring failure modes emerge, illustrated by the representative samples in Figure~\ref{fig:error_analysis}.

\textbf{(i) Informative content with aggressive formatting.} The first case shows a presenter holding a National ID card beneath large, colorful Bengali text stating that the card can be downloaded from home; the title (``How to download NID card $\vert$ National ID Download'') is plainly instructional, and the correct label is non-clickbait. The model nevertheless flags it: the visual encoder associates dense text overlays, bright colors, and the ``person holding an object'' layout with promotional clickbait, while the text encoder cannot separate marketing-style keywords (``Download'') from legitimate service language. The model lacks the contextual knowledge that government-service tutorials adopt visually striking designs for reach, not deception.

\textbf{(ii) Metaphorical news language.} The second case is a football training report titled ``Apart from India, another opponent for Bangladesh is Shillong's `weather'! $\vert$ BD Football'', genuine sports journalism using weather-as-opponent metaphor. The text encoder reads ``opponent'' and ``India'' as sensational rivalry cues and misses the metaphorical framing, illustrating the difficulty of figurative Bengali for current multilingual encoders.

\textbf{(iii) Polished visuals with cultural slang.} The third case pairs a professionally designed thumbnail of a Rafale fighter jet with the Prime Minister of India and the title ``How terrible is the Rafale fighter jet. Why is India jumping such strong about Rafale'', exaggerated, emotionally charged clickbait. The model predicts non-clickbait: the high production quality and the presence of a political figure read as documentary credibility, while the Bengali colloquialism ``lafalafi'' (excessive hype or fuss) loses its sensational tone under literal translation. The mismatch between polished visuals and provocative language defeats both encoders simultaneously.

These cases jointly indicate that future detectors need OCR over thumbnail text, richer modeling of Bengali figurative and colloquial language, and mechanisms that reason explicitly about \emph{intent} rather than surface style.

%%%%%%%%%%%%%%%%%%%%%%%%
%%%%%%%%%%%%%%%%%%%%%%%%
\section{Discussion}
\label{sec:discussion}

\subsection{Answers to the Research Questions}
\textbf{RQ1.} BanClickThumb demonstrates that a balanced, reliably annotated multimodal clickbait dataset can be constructed for Bengali at meaningful scale: 7,147 thumbnail-title pairs across five domains with inter-annotator agreement of 0.83 to 0.93. As Table~\ref{tab:comparative_simple} shows, it is the only resource that combines Bengali language coverage, manual annotation, and paired image-text modalities, complementing the much larger yet text-only BaitBuster-Bangla.

\textbf{RQ2.} Text is the stronger single modality: the best text model (XLM-RoBERTa, 0.82) outperforms the best vision model (SwiftFormer, 0.68) by 14 percentage points. Cross-lingually pre-trained XLM-RoBERTa consistently dominates mBERT and DistilBERT, confirming the value of large-scale multilingual pretraining for Bengali. Among vision backbones, the efficient SwiftFormer slightly outperforms Swin and ViT in isolation, suggesting that unimodal thumbnail classification benefits more from robust local feature extraction than from raw model capacity.

\textbf{RQ3.} Multimodal fusion helps, and fusion depth matters. All three strategies lift accuracy over the text-only baseline (early: 0.83; late: 0.83; intermediate: 0.84), with intermediate fusion, BanClickFusionFormer, performing best overall and most evenly across precision and recall. The gain over text alone is modest (2 points) yet the gain over vision alone is large (16 points), and the balanced error profile of intermediate fusion (Figure~\ref{fig:confusion_matrices}) indicates that thumbnails resolve cases that titles alone cannot. At the same time, the collapse of the intermediate Swin + XLM-RoBERTa pairing shows that intermediate fusion is sensitive to representation compatibility between backbones, whereas early and late fusion are more forgiving.

\textbf{RQ4.} The characteristic failures are not random noise yet systematic blind spots: dense legitimate text overlays mistaken for sensationalism, metaphorical journalism mistaken for rivalry-baiting, and culturally coded slang neutralized by translation. Each points to a concrete architectural remedy, OCR integration, figurative-language-aware encoders, and native-script modeling without translation, respectively.

\subsection{Theoretical Implications}
This study extends the literature on deceptive-content detection in three ways. First, it provides empirical evidence that clickbait in a low-resource language is a joint visual-linguistic construct: the complementarity of modalities observed here (weak vision-only performance, yet consistent fusion gains) supports theorizing clickbait as a \emph{promise-content mismatch} rather than a property of text alone. Second, the systematic comparison of fusion stages contributes to multimodal learning theory \citep{baltrusaitis2019multimodal} by showing that the advantage of intermediate fusion is conditional on encoder compatibility, a nuance rarely visible in single-configuration studies. Third, the error analysis surfaces the role of cultural and figurative language in deception detection, suggesting that transfer from high-resource benchmarks systematically underestimates the difficulty of low-resource, code-mixed environments.

\subsection{Practical Implications}
For platform operators and content moderators, the results indicate that a deployable Bengali clickbait screener can be built today from off-the-shelf transformers at 0.84 accuracy, with the efficient SwiftFormer + XLM-RoBERTa variant (0.83) offering a cheaper alternative for large-scale or on-device screening. For policymakers and media-literacy educators in Bangladesh's rapidly growing digital ecosystem, BanClickThumb provides an auditable, culturally grounded evidence base for characterizing manipulative content. For advertisers and brand-safety teams, automated thumbnail-title screening offers a mechanism to avoid placement alongside deceptive content. Finally, the public release of the dataset, code, and benchmarks lowers the entry barrier for researchers and practitioners building information-quality tools for the roughly 240 million Bengali speakers worldwide.

%%%%%%%%%%%%%%%%%%%%%%%%
%%%%%%%%%%%%%%%%%%%%%%%%
\section{Limitations and Future Research}
\label{sec:limitations}

Although BanClickThumb represents a meaningful step for multimodal Bengali clickbait detection, several limitations should be acknowledged, each of which charts a direction for future work.

\begin{itemize}
 \item \textbf{Platform and temporal scope.} The dataset is collected exclusively from YouTube videos published between March and October 2025. Because clickbait tactics evolve rapidly and differ across platforms, cross-platform transfer (e.g., Facebook, TikTok) and temporal robustness require domain adaptation studies on refreshed collections.
 \item \textbf{Binary labeling.} The clickbait/non-clickbait dichotomy does not capture the spectrum from mild exaggeration to deliberate deception. Finer-grained severity annotation would enable more nuanced training and evaluation.
 \item \textbf{Absence of OCR.} Text embedded within thumbnails, often the most manipulative element, is currently unmodeled. Integrating Bengali OCR into the visual pipeline is a priority, as the error analysis in Section~\ref{sec:error} directly implicates this gap.
 \item \textbf{Scale and class weighting.} At 7,147 samples the dataset is moderate in size, and class weighting was applied systematically only in the text branch; harmonizing imbalance handling across all modalities and fusion configurations is needed.
 \item \textbf{Statistical rigor.} Results are based on single training runs without confidence intervals or significance testing. Multiple seeded runs, convergence curves, and significance analysis would strengthen comparative conclusions.
 \item \textbf{Baselines and ablations.} Comparisons with pre-trained vision-language models such as CLIP \citep{radford2021learning} and BLIP \citep{li2022blip}, including zero-shot and few-shot evaluation, remain to be conducted, as do ablations isolating the contributions of the translation/transliteration module, individual preprocessing steps, fusion-layer dimensionality, and the late fusion weight $\alpha$.
 \item \textbf{Richer modalities and explainability.} Extending the framework to video frames and speech transcripts would allow verifying whether video content semantically fulfills its thumbnail and title, and Grad-CAM and token-level attention mapping would improve transparency for human moderators. A complementary quantitative error analysis across content domains would systematize the qualitative findings reported here.
\end{itemize}

%%%%%%%%%%%%%%%%%%%%%%%%
%%%%%%%%%%%%%%%%%%%%%%%%
\section{Conclusions}
\label{sec:conclusions}

In this paper, we introduced BanClickThumb, the first publicly available multimodal dataset for Bengali YouTube clickbait detection, comprising 7,147 thumbnail-title pairs annotated with high inter-annotator agreement. We established systematic transformer benchmarks across unimodal and multimodal settings: the text-only BanClickTextFormer (XLM-RoBERTa) achieved 0.82 accuracy, the image-only BanClickImageFormer (SwiftFormer) achieved 0.68, and the proposed intermediate fusion BanClickFusionFormer (ViT + XLM-RoBERTa) achieved the best accuracy of 0.84, an improvement of 2 percentage points over the text baseline and 16 percentage points over the visual baseline. Beyond the headline numbers, our comparison of 27 fusion configurations shows that intermediate fusion is the most effective and also the most sensitive to encoder compatibility, and our error analysis identifies dense thumbnail text, metaphorical language, and cultural slang as the frontier challenges. Although these results are encouraging, the achieved accuracy indicates substantial room for improvement through larger datasets, OCR-augmented pipelines, and pre-trained vision-language models. We hope BanClickThumb and these benchmarks will serve as a durable foundation for research on multimodal deceptive-content detection in under-resourced languages.

%%%%%%%%%%%%%%%%%%%%%%%%
%%%%%%%%%%%%%%%%%%%%%%%%
\section*{CRediT authorship contribution statement}
\textbf{Md. Ariful Islam:} Conceptualization, Methodology, Software, Data Curation, Writing - Original Draft, Visualization, Validation. \textbf{Md. Tanvirul Islam:} Methodology, Software, Data Curation, Writing - Original Draft, Visualization, Validation. \textbf{Md. Maruf Hossain Miru:} Methodology, Data Curation, Writing - Original Draft, Visualization. \textbf{Md. Khalid Syfullah:} Supervision, Project Administration, Writing - Review \& Editing.

%%%%%%%%%%%%%%%%%%%%%%%%
%%%%%%%%%%%%%%%%%%%%%%%%
\section*{Declaration of competing interest}
The authors declare that they have no known competing financial interests or personal relationships that could have appeared to influence the work reported in this paper.

%%%%%%%%%%%%%%%%%%%%%%%%
%%%%%%%%%%%%%%%%%%%%%%%%
\section*{Data availability}
The BanClickThumb dataset is publicly available on Kaggle: \url{https://kaggle.com/datasets/5345efaf68005de56157e10096c251f1714d5acc19bd85857275ef11be2d5fab}.

%%%%%%%%%%%%%%%%%%%%%%%%
%%%%%%%%%%%%%%%%%%%%%%%%
\section*{Code availability}
The code and supplementary files are publicly available on GitHub: \url{https://github.com/arifbaust10/BanClickThumb-A-Multimodal-Dataset-and-Benchmarks-for-Clickbait-Detection-in-Bengali-YouTube-Videos}.

%%%%%%%%%%%%%%%%%%%%%%%%
\section*{ Declaration of generative AI and AI-assisted technologies in the manuscript preparation process}

During the preparation of this manuscript, the authors used generative AI and AI-assisted technologies to support language refinement, improve the clarity and organization of the manuscript, and assist in proofreading during the writing process. The use of these tools was limited to editorial and linguistic assistance. The authors critically reviewed, verified, and edited all AI-assisted outputs to ensure the accuracy, originality, and integrity of the manuscript. The scientific content, methodology, experimental design, analysis, interpretation of results, and conclusions presented in this work are solely the responsibility of the authors. The authors take full responsibility for the content of the published article.

%%%%%%%%%%%%%%%%%%%%%%%%
\bibliographystyle{elsarticle-harv}
\bibliography{references}

%%%%%%%%%%%%%%%%%%%%%%%%
\end{document}